\documentclass{article}

\usepackage{arxiv}

\usepackage[utf8]{inputenc} 
\usepackage[T1]{fontenc}    
\usepackage{hyperref}       
\usepackage{url}            
\usepackage{booktabs}       
\usepackage{amsfonts}       
\usepackage{nicefrac}       
\usepackage{microtype}      
\usepackage{lipsum}
\usepackage{graphicx}
\graphicspath{ {./images/} }

\usepackage{multirow}
\usepackage{diagbox}
\newcommand{\etal}{et.~al.~}
\usepackage{amsmath}
\usepackage{amssymb}
\usepackage{booktabs}

\title{Multi-Scale Relational Graph Convolutional Network for Multiple Instance Learning in Histopathology Images}

\author{
 Roozbeh Bazargani \\
  The University of British Columbia\\
  Vancouver, BC V6T 1Z4 \\
  \texttt{roozbehb@ece.ubc.ca} \\
   \And
 Ladan Fazli \\
  The Vancouver Prostate Centre\\
  The University of British Columbia\\
  Vancouver, BC V6T 1Z4 \\
  \texttt{ladan.fazli@vch.ca} \\
  \And
 Larry Goldenberg \\
  The Vancouver Prostate Centre\\
  The University of British Columbia\\
  Vancouver, BC V6T 1Z4 \\
  \texttt{l.gold@ubc.ca} \\
  \And
  Martin Gleave \\
  The Vancouver Prostate Centre\\
  The University of British Columbia\\
  Vancouver, BC V6T 1Z4 \\
  \texttt{m.gleave@ubc.ca} \\
   \And
 Ali Bashashati$^*$ \\
  The University of British Columbia\\
  Vancouver, BC V6T 1Z4 \\
  \texttt{ali.bashashati@ubc.ca} \\
  \And
 Septimiu Salcudean$^*$ \\
  The University of British Columbia\\
  Vancouver, BC V6T 1Z4 \\
  \texttt{tims@ece.ubc.ca} \\
}

\begin{document}
\maketitle
\def\thefootnote{*}\footnotetext{Co-senior authors}

\begin{abstract}
Graph convolutional neural networks have shown significant potential in natural and histopathology images. However, their use has only been studied in a single magnification or multi-magnification with late fusion. In order to leverage the multi-magnification information and early fusion with graph convolutional networks, we handle different embedding spaces at each magnification by introducing the Multi-Scale Relational Graph Convolutional Network (MS-RGCN) as a multiple instance learning method. We model histopathology image patches and their relation with neighboring patches and patches at other scales (i.e., magnifications) as a graph. To pass the information between different magnification embedding spaces, we define separate message-passing neural networks based on the node and edge type. We experiment on prostate cancer histopathology images to predict the grade groups based on the extracted features from patches. We also compare our MS-RGCN with multiple state-of-the-art methods with evaluations on several source and held-out datasets. Our method outperforms the state-of-the-art on all of the datasets and image types consisting of tissue microarrays, whole-mount slide regions, and whole-slide images. Through an ablation study, we test and show the value of the pertinent design features of the MS-RGCN.
\end{abstract}


\section{Introduction}
\label{sec:intro}

\begin{figure*}[!t]
  \centering
  \includegraphics[width=\linewidth]{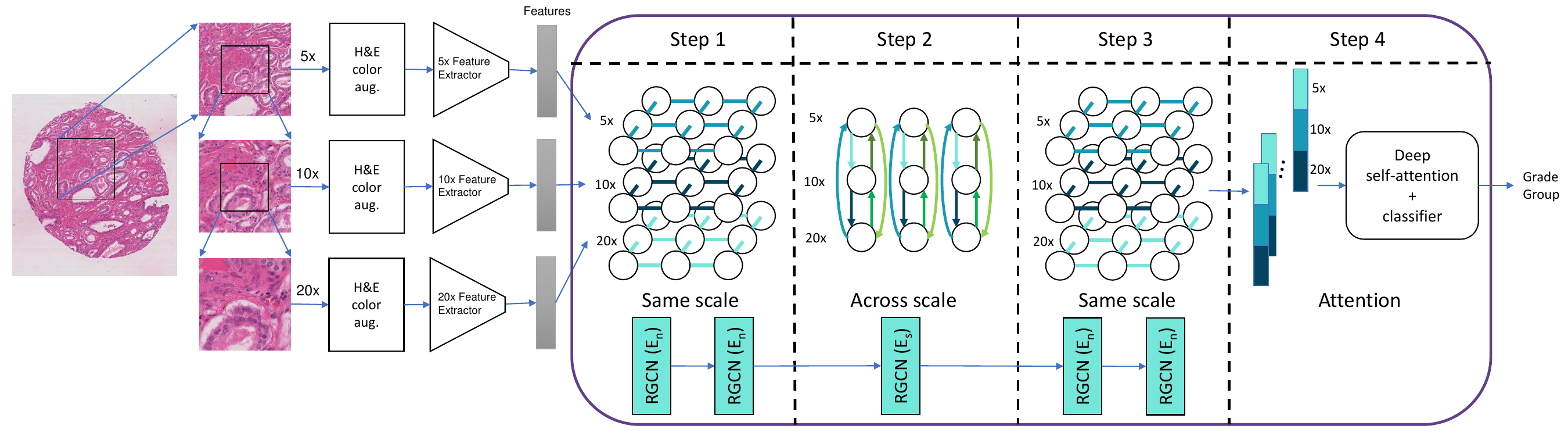}
  \caption{Overview of the model. Patch extractions at $5\times$, $10\times$, and $20\times$ were carried out with higher resolution patches being at the center of the prior resolution. 
We use a combination of stain-color and color augmentation to improve the performance and generalizability of the models to the held-out datasets. The feature extractors were trained on predicting the patch-level annotations in order to extract features. A Graph based on the patches is constructed where each node represents a patch and different edge types exist based on the relation to either neighboring or across magnification patches. Our novel method utilizes these different edge relations in four steps: 1) Two layers of RGCN have been used between neighboring nodes at each scale to get more robust features at each location by using the surrounding features; 2) One layer of the RGCN across magnification edges translates the features of each magnification to the other ones at each location; 3) Two RGCN layers on the neighboring edges aim to combine and reduce features for the final prediction, and finally; 4) A deep self-attention to better attend to the complex features and a two-layer fully-connected neural network to output the final image-level label.}
  \label{fig:fig-graph}
\end{figure*}

Multiple Instance Learning (MIL) is a weakly-supervised learning method that is widely utilized to train classification models with image-level annotations while detailed patch-level or pixel-level annotations are not available. 
Over the past years, a variety of MIL-based techniques utilizing various attention mechanisms, transformers, and Graph Neural Network (GNN) architectures have been introduced for image-based classification. Moreover, MIL-based techniques have been widely utilized in the context of giga-pixel histopathology image analysis \cite{zhao2020predicting, chen2022scaling, alon2022neuroplastic, marini2022unleashing, adnan2020representation}.

Computer-aided classification of histopathology images compared to natural images involves tackling several challenges as follows: 

1) The large size of the images (typically $100,000$ $\times$ $100,000$ pixels) makes it difficult to utilize off-the-shelf vision models. Therefore, conventional histopathology classification models rely on the extraction of smaller images, or {\em patches}, from the full image typically examined by the pathologist. Afterward, the predictions for individual patches are aggregated to achieve image-level predictions.

2) While most existing image classification models take, as input, images at a fixed magnification, important salient morphological features at different optical zooms are in fact utilized by pathologists for accurate disease diagnosis and classification. 
Therefore, models are required to capture features such as cells and nuclei size, as well as high-level features such as context and tissue structure for improved classification. 

3) While classification methods require labeled data for training, acquiring detailed cell-level annotations for histopathology is known to be a complex and tedious task that requires years of training. Furthermore, the complexity of annotations results in inter-observer variability among pathologists, and even amongst experts \cite{nir2018automatic}.

4) The performance of many existing models suffers when there is a domain shift between the training and testing data. In such scenarios, models trained on the data from one center may not perform well on data from other centers. This is widely observed in histopathology as even standard hematoxylin and eosin staining produce large color variations. 
Color normalization, color augmentation, and Domain Adversarial Neural Network (DANN) are among the methods that have been used to improve the generalizability of the classification methods~\cite{vahadane2016structure, tellez2019quantifying, hashimoto2020multi}.

To tackle the time-consuming task of performing detailed annotations by pathologists, 
MIL has gained substantial attention as a weakly-supervised learning method.
Most of the State-Of-The-Art (SOTA) papers use attention-based MIL, either at a single magnification~\cite{ilse2018attention, lu2021data} or multi-scale~\cite{li2021dual, thandiackal2022differentiable, yao2020whole}. The latter is done in order to better capture the analysis procedure that pathologists have and let the model look at a larger area of the image at lower magnifications while having a high resolution at higher magnifications. 
Since graphs efficiently describe the relation between tissue regions and patches, graph-based deep learning has shown promising results in computational histopathology in grading and survival analysis \cite{ahmedt2021survey, li2018graph}. 
Zhao \etal \cite{zhao2020predicting} used Deep Graph Convolutional Network (DGCN) as a MIL method at a single magnification that unveiled the potential of GNNs in histopathology. There are two studies of multi-scale graph convolutional networks that have been just published. Zhang \etal \cite{zhang2022ms} used Graph Convolutional Networks (GCNs) at each magnification and summed the GCN outputs at the end for classification, and Alon \etal \cite{alon2022neuroplastic} used a multi-scale graph attention network for nuclei segmentation. Recently, relational and heterogeneous GNNs that use different edge types and assign them separate message-passing neural networks~\cite{schlichtkrull2018modeling, zhang2019heterogeneous} have drawn much attention. Their potential in computer vision \cite{wang2020contextual, han2021query}, motivated its use in this work, for a better representation of the relation between patches at different magnifications.

In this paper, 
we introduce the Multi-Scale Relational Graph Convolutional Network (MS-RGCN) for MIL-based aggregation of multi-scale patches in histopathology images with different node and edge types as detailed in Figure~\ref{fig:fig-graph}. We experimented with Vision Transformers (ViT) and Convolutional Neural Network (CNN) feature extractors for the classification of prostate cancer (PCa) in Tissue Micro Arrays (TMAs) and selected the superior approach for the evaluation of the model on {\em Whole Slide Images} (WSIs). PCa classification is known to be challenging with high inter-observer classification variability amongst pathologists and therefore, suitable for evaluation of our proposed model. 

Our proposed method handles the first two challenging aspects of histopathology image analysis and is evaluated under the other two as outlined earlier: 1) by passing information between patches, we are able to see a large section of the image while extracting only small patches, 2) by using different edge and node types within and between different magnifications, we preserve the various features at each magnification that might contribute to better diagnostic performance, 3) only a small part of the data is annotated and has been used to train the feature extractor from it, and 4) the models are trained on a tissue microarray (TMA) dataset from one center and tested on three diverse datasets representing biopsy and radical prostatectomy whole section slides and TMAs from Netherlands and Sweden \cite{bulten2022artificial}, Switzerland \cite{arvaniti2018automated}, and United States. 


The main contributions of this work include:
\begin{itemize}
    \item First work in histopathology to design relational graphs and several node and edge types to tackle the problem of different embedding spaces at each magnification; showing the importance of representing pathologists' analysis procedure.
    \item Extensive comparison with eight SOTA models with ViT and CNN feature extractors on three held-out datasets representing more than 7000 TMA cores and slides to demonstrate the importance of multi-magnification and usage of relational graphs for generalization.
    \item Qualitative analysis of the explainability of the model by comparing self-attention heatmaps with pathologist masks.
\end{itemize}

\section{Related Work}

\subsection{Multiple Instance Learning methods}
Zaheer and Brendel \etal \cite{zaheer2017deep, brendel2019approximating} used bags of local features, operating on sets as a MIL task, and showed improvement on natural images. Ilse \etal \cite{ilse2018attention} was the first attention-based MIL in histopathology images. Since then, different types of MIL and aggregation of patches have been investigated on histopathology images \cite{tellez2019neural, yao2019deep, yao2020whole, chen2021multimodal, lerousseau2021sparseconvmil, schirris2022deepsmile, hou2016patch, carmichael2022incorporating}. Moreover, multi-scale methods have been shown to improve segmentation and classification by integrating high resolution and high field-of-view from different scales \cite{li2021dual, zhang2022ms, hashimoto2020multi}.

\subsubsection{Attention-based MIL.} DeepMIL by Ilse \etal used two layers of fully connected neural networks to predict the attention \cite{ilse2018attention}. Clustering-Constrained-Attention MIL (CLAM) by Lu \etal trained attention branches using clustering losses \cite{lu2021data}. Dual-Stream-MIL (DS-MIL) by Li \etal concatenated features of $5\times$ and $20\times$ magnifications and trained a MIL aggregator on the concatenated feature vectors \cite{li2021dual}. ZoomMIL by Thandiackal \etal used attention at each magnification then summed the aggregation from each magnification and passed the results to a classifier \cite{thandiackal2022differentiable}.

\subsubsection{Transformer-based MIL.} TransMIL by Shao \etal \cite{shao2021transmil} uses a transformer-based correlated MIL based on Nystromformer that has a nystrom-based self-attention \cite{xiong2021nystromformer}. Hierarchical Image Pyramid Transformer (HIPT) by Chen \etal \cite{chen2022scaling} uses three Vision Transformers on WSIs, ViT$_{256-16}$, ViT$_{4096-256}$, and ViT$_{\text{WSI}-4096}$, separately with self-supervised tasks in ViT$_{256-16}$ and ViT$_{4096-256}$, where ViT$_{b-a}$ denotes a ViT that takes $a\times a$ pixel images and outputs the final feature vector of $b \times b$ pixel image after combining its features. This approach used cell, patch, region, and whole image information at each of the $16$, $256$, $4096$, and WSI levels, respectively.

\subsubsection{Graph-based MIL.} DGCN by Zhao \etal \cite{zhao2020predicting} used a Variational AutoEncoder and Generative Adversarial Network (VAE-GAN) as a feature extractor, with three GCNs with a Self-Attention Graph Pooling (SAGPooling) \cite{lee2019self} and two fully-connected layers as a MIL network at the end. They used two layers of fully-connected neural networks as a message passer in their three GCN layers. Moreover, they added edges between two patches if their distance was less than half of the maximum distance between patches. A Multi-Scale Graph Wavelet Neural Network (MS-GWNN) by Zhang \etal \cite{zhang2022ms} trained a late-fusion aggregation of multi-scale GNNs where they had a graph for each magnification separately and then summed the predictions of the graphs at the end.

\section{Material and methods}
We chose to carry out our experiments on prostate cancer.
PCa is the second most common cancer in men~\cite{siegel2019cancer} and the fifth leading cause of death worldwide~\cite{rawla2019epidemiology}. PCa is
a heterogeneous disease, with a diverse range of dissimilar histological patterns with the same severity score, making their classification and risk stratification challenging \cite{nir2018automatic}. This is demonstrated by the large inter-observer variability among pathologists with overall unweighted kappa ($\kappa$) coefficient of $0.435$ for general pathologists \cite{allsbrook2001interobserver1} and overall weighted $\kappa$ range of $0.56$-$0.70$ for urologic pathologists \cite{allsbrook2001interobserver2}. This leads to under- or over-treatment of patients and thus impacts their survival rate, quality of life, and healthcare system costs~\cite{berney2014reasons}.

Histological patterns and/or sub-patterns are characteristic of particular tumors or groups of PCa tumors~\cite{dive2014histological}. 
The severity of morphological changes seen in the Hematoxylin and Eosin (H\&E) stained tissue is graded as Gleason Patterns (GP), with severity ranging from 1 to 5. The Gleason score (GS) is based on the most and second most dominant patterns. Based on GS, there are 5 groups, in which group 2 is considered low risk, and groups 4, and 5 are considered high-grade cancerous tissue~\cite{pierorazio2013prognostic}. Epstein \etal \cite{epstein20162014} introduced 5 grade groups (GG) based on GSs as follows: GG1) GS $\leq 6$, GG2) GS3+4=7, GG3) GS4+3=7, GG4) GS8, and GG5) GS9 or GS10.

We used prostate cancer datasets from different centers to evaluate our proposed model on TMAs (from Canada and Switzerland), Whole Mount Slide (WMS) regions (from the United States), and WSI biopsy datasets (from Netherlands and Sweden). The dataset from Canada consists of 1082 TMA cores from 493 radical prostatectomy patients from the Vancouver Prostate Center that were digitized at $40\times$ magnification using an SCN400 Slide Scanner (Leica Microsystems, Wetzlar, Germany). A subset of 333 TMA cores (231 patients) from the Vancouver dataset was annotated by six pathologists with different levels of experience. This subset is used for training feature extractors for all of the experiments. The majority vote was used to calculate the final annotation mask \cite{nir2018automatic,karimi2019deep}. We used another TMA dataset from Zurich (Switzerland) to test our model. This dataset consists of 886 TMA cores at $40\times$ magnification from 886 patients~\cite{arvaniti2018automated}. We also used a private dataset from the University of Colorado School of Medicine (United States) for testing, consisting of 230 WMS from 56 patients who underwent radical prostatectomy. The WMSs were digitized at $20\times$ magnification using an Aperio ScanScope XT (Leica Biosystems, Vista, CA, USA). The Gleason patterns were annotated by the consensus of five pathologists. We randomly extracted 1116 regions from WMSs of approximately $2560\times2560$ pixels for classification and labeled them with a GG based on the GP maps in that region. Finally, to evaluate the performance of our model on WSIs and biopsy material, we utilized the publicly available portion of the PANDA challenge dataset \cite{bulten2022artificial} digitized at $20\times$ magnification. This dataset consists of two centers, Radboud (Netherlands) and Karolinska (Sweden).  Table \ref{tab:data} shows the GG label distribution in the datasets.

\begin{table}[!t]
\centering
\begin{tabular}{ccccccc|c}
\hline
Dataset & Benign & GG1 & GG2 & GG3 & GG4 & GG5 & Total \\ \hline
Vancouver & 221 & 320 & 217 & 126 & 131 & 67 & 1082 \\
Zurich & 115 & 277 & 85 & 50 & 221 & 138 & 886 \\
Colorado & 276 & 311 & 106 & 75 & 262 & 86 & 1116 \\
Karolinska & 1924 & 1810 & 666 & 317 & 480 & 250 & 5447 \\
Radboud & 948 & 801 & 673 & 908 & 764 & 963 & 5057
\end{tabular}%
\caption{Data distribution in the datasets.}
\label{tab:data}
\end{table}


\subsection{Patch extraction}
For patch extraction, we extracted $256\times256$ patches in a non-overlapping manner at the highest resolution, in our case $20\times$. At the lower magnifications ($10\times$, and $5\times$), we extracted $256\times256$ patches in a way that the respective patch in the higher resolution would be in the center as demonstrated in Figure \ref{fig:fig-graph}. Therefore, the lower resolution patches might go out of the image bounds at the edges of the Tissue Micro Array (TMA) image where we have padded them with white pixels (value of $255$ in all of the three channels). To construct the graph, the set of vertices $V$ is computed as $V = V_{5\times} \cup V_{10\times} \cup V_{20\times}$ where $V_{m\times}$ is the set of patches from the whole image at magnification $m$. The set of edges is $E = E_n \cup E_s$
where $E_n$ represents edges between the neighboring patches (up, down, left, and right at one magnification) and $E_s$ represents edges between the same location patches across magnifications. We used the neighboring patches to locally improve the extracted features. Finally, we define the graph of the whole image as $G=(V,E)$. In Figure \ref{fig:fig-graph}, the manner of constructing a graph for the whole image is demonstrated.

\subsection{Feature extractor}
We used both ViT \cite{dosovitskiy2020image} and CNN based feature extractors in our experiments with TMAs to show the capability of our model with both feature extractor types. Then, we used the superior one for testing on WMSs and WSIs. We use transfer learning on the small set of patches from the Vancouver dataset with their patch-level annotations \cite{nir2018automatic, karimi2019deep}. We choose ViT\_b\_16 \cite{dosovitskiy2020image} for our ViT model to obtain cell-level information at $16\times16$ pixels images \cite{chen2022scaling} ($6.88\times6.88 \mu$m$^2$ resolution at $20\times$ magnification) from $256\times256$ pixels patches extracted at each magnification. Due to a large number of parameters in ViT\_b\_16, only the last encoder and classifier of the pretrained model on the {\em ImageNet} dataset \cite{russakovsky2015imagenet} are further trained with patch-labels for the patch-based classifications. Using this set of classes, the feature extractor would learn features that correspond to the image-level labels and therefore result in a better image-level classification in the final output. Moreover, for the CNN model we select ResNet18 \cite{he2016deep} and updated all of the weights of the pretrained model on {\em ImageNet} dataset by performing patch-based classification. For both ViT and CNN models, we used the weighted cross entropy loss function to tackle the imbalanced classes in the datasets. Furthermore, to improve model generalizability on held-out sets, we utilize a combination of color normalization and augmentation \cite{bazargani2023novel}, that was inspired by~\cite{tellez2019quantifying, boschman2022utility}.

\subsection{Graph-based MIL}
Previously, only Zhang \etal \cite{zhang2022ms} used GNNs as a multi-scale aggregator. However, they have not used any edges between different magnifications and used three separate GNNs at each magnification with a sum at the end as a late-fusion of the information. In contrast, we use edges between magnifications in order to provide an early fusion of information between different magnifications. This follows Dwivedi \etal \cite{dwivedi2022multi}, which showed that early fusion of information helped the model perform better than late fusion.

The main difficulty in combining features from different magnifications is that morphological features at different magnifications have different interpretations.
To solve this issue, after constructing our graph, we gave each node a type based on the magnification the node belongs to. So the set of types $T$ that a vertex (node) $v \in V$ belongs to consists of three elements $T = \{5, 10, 20\}$. We used different edge types for connecting different types of nodes. The set of types $R$ that an edge $e$ belongs to is the Cartesian product of $T$ with itself
\begin{equation}
\begin{aligned}
    R = T \times T = \{ (5,5), (10,10), (20,20), (5,10), \\ (10, 5), (5,20), (20,5), (10,20), (20,10) \}
\end{aligned}
\end{equation}

where $(a,b)$ indicates the type of edge that goes from a vertex of type $a$ to a vertex of type $b$. We partition the set of edge types into neighboring edge types, and scaling edge types as $R_n = \{ (5,5), (10,10), (20,20) \}$ on $E_n$ and $R_s = \{ (5,10), (10, 5), (5,20), (20,5), (10,20), (20,10) \}$ on $E_s$, respectively, where $R_n \cap R_s = \emptyset$ and $R_n \cup R_s = R$. As illustrated in Figure \ref{fig:fig-graph}, our method consists of four sections based on the sets $R_n$ and $R_s$, as discussed next.

At the first step, we used two layers of RGCN \cite{schlichtkrull2018modeling}, based on Graph Convolutional Networks (GCN) \cite{kipf2016semi}. In a RGCN, we have 
\begin{equation} \label{eq:rgcn}
    h^{\prime}_i = b + W_{root} \cdot h_i + \frac{1}{|R_k|}\sum_{r \in R_k} \sum_{j \in N_r(i)} \frac{1}{|N_r(i)|} W_r \cdot h_j,
\end{equation}
where $h_i$, and $h^\prime_i$ are the current and output embeddings of node $i$, $b$ is the bias vector, $R_k$ is the edge type, which can be either $R_n$ or $R_s$, $N_r(i)$ is the set of neighboring nodes of node $i$ based on the edges of type $r$, and $W_{root}$ and $W_r$ are trainable weights that are applied to the node itself and neighboring nodes based on edge type $r$, respectively. We applied these layers on the $R_n$ edges to make the extracted features more robust by a learned weighted averaging based on the features of neighboring patches. This will let the network see features of neighboring patches at the distance of 2 vertices in the graph which are 12 vertices. We do not use non-linear activation functions here, as we want linear mapping of the features that we already have in order to see the greater distance without changing the features. To experiment the benefit, in the ablation study \ref{sec:ablation} there is a case where we have used a non-linear activation function in this step.

Now, having more robust features at each scale, we interpret the features of other scales to the features of the scale where the node belongs to in the message passing network. For this purpose, we used one layer of RGCN on $R_s$ edges with a layer-normalization \cite{ba2016layer} and Rectified Linear Unit (ReLU) activation function. Different edge types and as a result, different message passing networks will let the model effectively pass the information between different node types. For instance, the network on edge type $(10,20)$ learns how to translate the features at $10\times$ to features at $20\times$. An example of how a node at $20\times$ will receive the information from other magnifications is based on Equation \ref{eq:rgcn}. Information from $5\times$ and $10\times$ will be translated to $20\times$ via $(5,20)$ and $(10,20)$ edges, respectively. Then, the aggregation of the translated information and the information from the $20\times$ vertex will replace the node features. This multi-scale idea lets each node have a broad field of view at a higher resolution. For example, a node at $20\times$ can see 13 patches at $5\times$ and vice versa. Moreover, it is important to note that at lower magnifications it is much easier to distinguish benign from cancerous regions, and at higher magnifications it is easier to distinguish the subtypes from each other.

Next, we start to change the features in each magnification using two layers of RGCN on $R_n$ edges with layer-normalization and ReLU activation function. It is important to note that each magnification has different features so we can not combine them with GCN layers instead of RGCN layers. This is why after RGCN layers we concatenate the features of different scale nodes at each location instead of averaging them. Next, we carry out our final step which is pooling this bag of nodes.

Finally, each location in the main image has a concatenated feature vector. Considering the complexity of the histopathology images and their features, we use the pooling operation from Ilse \etal \cite{ilse2018attention}, where the attention values are learned by a neural network. After pooling, a two-layer fully-connected classifier was trained to recognize the image-level label.

\section{Results and Discussion}

The feature extractors of our proposed workflow were trained with a batch size of $32$ patches of size $256\times256$. The classes consist of {\em Background, Benign, GP3, GP4, GP5}, and {\em Others} that includes stroma or non-tumor regions.
We used the Adam optimizer \cite{kingma2014adam} with learning rate of $0.001$, $\beta_1 = 0.9$, and $\beta_2 = 0.999$. In the end, we extracted embeddings of each patch at the final encoder layer of ViT\_b\_16 and average pooling layer of ResNet18. It is of note to mention that we trained feature extractors only on the Vancouver dataset. We also trained two separate MIL models, one for TMAs and WMS regions and one for WSIs. The MIL method for TMAs was trained based on the Vancouver dataset. For WSIs, we utilized the Karolinska dataset for MIL training and tested it on Radboud as the class distribution for the Radboud dataset is more uniform and has GP annotations for model evaluations. WSIs unlike TMA images are larger and have more patches, therefore, necessitating the need to train another MIL for these datasets. 

To compare the different MIL methods, we used the same feature extractor and froze its weights in all of the experiments. The task for training MILs was to classify the whole image (TMA, WMS, or WSI) into 6 classes of Benign (BN), and GG1-5. The batch size of $32$ images, and Adam optimizer with a learning rate of $0.001$, $\beta_1 = 0.9$, and $\beta_2 = 0.999$ were used for training. The scheduler of the learning rate was \textit{ReduceLROnPlateau} with factor $= 0.3$, and patience $= 10$. 

We used 5-fold cross-validation training to eliminate selection bias and split the TMA cores into training, validation, and test sets based on patients and slides in order to have no overlap in feature extractor and MIL methods training. We used two different criteria: macro-averaged Area Under the Receiver Operating Characteristic Curve (AUC), and quadratic-weighted Cohen's $\kappa$ \cite{cohen1960coefficient} to compare the performance of models. 
Cohen's $\kappa$ has been used on histopathology to determine the agreement among pathologists and artificial intelligence models based on the confusion matrix \cite{strom2020artificial}. For computing quadratic-weighted-$\kappa$, we weigh diagonal elements of the confusion matrix $0$, and weigh the rest of the elements based on its quadratic distance. As a result, we would penalize the prediction of ground truth of GG2 as GG5 more than predicting it as GG3. AUC can take a value between 0 and 1 where $0.5$ means random prediction, and $\kappa$ can take a value ranging from -1 to 1 where between -1 and 0 the agreement is by chance and 1 means complete agreement. Thus, higher AUC and higher $\kappa$ are better.

\subsection{Comparison with state-of-the-art}

\begin{table*}[!t]
\centering
\caption{Comparison with state-of-the-art MIL methods with ViT feature extractors. We report average and standard deviation of macro-averaged AUC and quadratic weighted $\kappa$ on 5-folds. The best performance is shown in {\bf bold}.
}
\begin{tabular}{ccccc}
\hline
\multirow{2}{*}{Method} & \multicolumn{2}{c}{{Vancouver dataset}} & \multicolumn{2}{c}{{Zurich dataset}} \\ \cline{2-5}
& kappa & AUC (\%) & kappa & AUC (\%) \\ \hline
DeepMIL \cite{ilse2018attention} & $0.665 \pm 0.030$ &$82.57 \pm 1.95$ & $0.671 \pm 0.083$& $80.92 \pm 2.66$ \\
{CLAM-SB \cite{lu2021data}} & $0.647 \pm 0.026$ &$82.45 \pm 2.70$ & $0.674 \pm 0.062$& $80.62 \pm 2.25$  \\
DGCN \cite{zhao2020predicting} & $0.623 \pm 0.055$ &$77.09 \pm 2.18$ & $0.516 \pm 0.088$& $77.18 \pm 1.17$ \\
{HIPT-MIL \cite{chen2022scaling}} & $0.647 \pm 0.017$ &$81.28 \pm 1.87$ & $0.664 \pm 0.045$& $80.83 \pm 1.48$ \\
{TransMIL \cite{shao2021transmil}} & $0.634 \pm 0.068$ &$80.22 \pm 4.28$ & $0.576 \pm 0.085$& $78.36 \pm 1.48$ \\
{DS-MIL \cite{li2021dual}} & $0.493 \pm 0.154$ &$77.56 \pm 4.15$ & $0.659 \pm 0.035$& $78.99 \pm 2.20$ \\
{ZoomMIL \cite{thandiackal2022differentiable}} & $0.626 \pm 0.043$ &$79.29 \pm 3.49$ & $0.714 \pm 0.049$& $81.17 \pm 1.96$ \\
{MS-GWNN \cite{zhang2022ms}} & $0.549 \pm 0.061$ &$82.00 \pm 0.72$ & $0.710 \pm 0.017$& $80.32 \pm 0.93$ \\ \hline
{MS-RGCN (ours)} & $\mathbf{0.712 \pm 0.061}$ &$\mathbf{83.97 \pm 1.64}$ & $\mathbf{0.765 \pm 0.020}$& $\mathbf{83.12 \pm 0.60}$ \\ \hline
\end{tabular}%
\label{tab:sota}
\end{table*}

\begin{table*}[!t]
\centering
\caption{Comparison with state-of-the-art MIL methods with ResNet18 feature extractors. We report average and standard deviation of macro-averaged AUC and quadratic weighted $\kappa$ on 5-folds. The best performance is shown in {\bf bold}.
}
\resizebox{\textwidth}{!}{%
\begin{tabular}{ccccccccc}
\hline
 \multirow{2}{*}{Method} & \multicolumn{2}{c}{{Vancouver dataset}} & \multicolumn{2}{c}{{Zurich dataset}} & \multicolumn{2}{c}{{Colorado dataset}} & \multicolumn{2}{c}{{Radboud dataset}} \\ \cline{2-9}
 & kappa & AUC (\%) & kappa & AUC (\%) & kappa & AUC (\%) & kappa & AUC (\%) \\ \hline
DeepMIL \cite{ilse2018attention} & $ 0.714 \pm 0.019 $ &$ 84.39 \pm 1.44 $ & $ 0.734 \pm 0.029 $ & $ 82.64 \pm 1.59 $ & $ 0.580 \pm 0.052 $& $ 78.84 \pm 0.78 $ & $ 0.388 \pm 0.075 $& $ 68.60 \pm 4.92 $ \\
{CLAM-SB \cite{lu2021data}} & $ 0.672 \pm 0.050 $ &$ 84.11 \pm 1.65 $ & $ 0.735 \pm 0.007 $ & $ 82.18 \pm 0.95 $ & $ 0.570 \pm 0.047 $& $ 78.67 \pm 1.72 $ & $ 0.420 \pm 0.070 $& $ 70.84 \pm 4.18 $ \\
DGCN \cite{zhao2020predicting} & $ 0.640 \pm 0.068 $ &$ 80.62 \pm 1.45 $ & $ 0.638 \pm 0.095$ & $ 78.81 \pm 1.63 $ & $ 0.561 \pm 0.035 $& $ 75.45 \pm 1.35 $ & $ 0.412 \pm 0.089 $& $ 69.76 \pm 2.82 $ \\
{HIPT-MIL \cite{chen2022scaling}} & $ 0.666 \pm 0.044 $ &$ 83.52 \pm 1.37 $ & $ 0.724 \pm 0.013$ & $ 81.56 \pm 1.20 $ & $ 0.576 \pm 0.033 $& $ 78.47 \pm 2.08 $ & $ 0.457 \pm 0.060 $& $ 71.14 \pm 2.32 $ \\
{TransMIL \cite{shao2021transmil}} & $ 0.603 \pm 0.086 $ &$ 81.75 \pm 1.74 $ & $ 0.679 \pm 0.013$ & $ 81.41 \pm 1.17 $ & $ 0.514 \pm 0.117 $& $ 76.19 \pm 4.31 $ & $ 0.366 \pm 0.052 $& $ 69.89 \pm 2.90 $ \\
{DS-MIL \cite{li2021dual}} & $ 0.618 \pm 0.049 $ &$80.72 \pm 1.71 $ & $ 0.710 \pm 0.040$ & $ 81.93 \pm 2.02 $ & $ 0.597 \pm 0.034 $& $78.13 \pm 0.61 $ & $ 0.531 \pm 0.036 $& $76.34 \pm 2.57 $ \\
{ZoomMIL \cite{thandiackal2022differentiable}} & $ 0.689 \pm 0.065 $ &$81.76 \pm 1.41 $ & $ 0.738 \pm 0.022$ & $ 83.62 \pm 0.87 $ & $ 0.601 \pm 0.031 $& $78.99 \pm 1.54 $ & $ 0.550 \pm 0.040 $& $75.94 \pm 1.86 $ \\
{MS-GWNN \cite{zhang2022ms}} & $ 0.612 \pm 0.101 $ &$82.55 \pm 1.17 $ & $ 0.735 \pm 0.024$ & $ 82.16 \pm 1.23 $ & $ 0.604 \pm 0.034 $& $79.25 \pm 1.81 $ & $ 0.374 \pm 0.102 $& $74.91 \pm 1.35 $ \\ \hline
{MS-RGCN (ours)} & $\mathbf{0.727 \pm 0.028}$ &$\mathbf{85.18 \pm 1.45}$ & $\mathbf{0.781 \pm 0.013}$ & $\mathbf{85.24 \pm 1.07}$ & $\mathbf{0.668 \pm 0.031}$& $\mathbf{80.86 \pm 0.89}$ & $\mathbf{0.577 \pm 0.025}$& $\mathbf{78.84 \pm 2.73}$ \\ \hline
\end{tabular}%
}
\label{tab:sota_res}
\end{table*}

We compared our method with SOTA models in single and multiple magnifications. For single magnification models, we evaluated the aggregators of DeepMIL, CLAM-SB, DGCN, HIPT-MIL, and TransMIL. DeepMIL and CLAM-SB are attention-based, DGCN is a graph-based, and HIPT-MIL and TransMIL are transformer-based representative models. For multi-magnification models, we compared our model with DS-MIL, and ZoomMIL as attention-based and MS-GWNN as a Graph-based MIL model.

The results of comparing our model with SOTA models are shown in Tables \ref{tab:sota} and \ref{tab:sota_res}. It can be seen that MS-RGCN is outperforming all of the models in ViT and ResNet18 feature extractors. MS-RGCN improves SOTA models performance by up to $21.9\%$ in quadratic weighted Cohen's $\kappa$, and $6.88\%$ in macro averaged AUC on the source domain and up to $24.9\%$ in $\kappa$ and $6.43\%$ in AUC on the Zurich set. We believe ResNet18 is working better because all of its weights have been updated during training. As a result, ResNet18 is capable of better extracting morphological features. We can see that DeepMIL has the closest performance to MS-RGCN on the source dataset. However, its performance drops significantly on the Zurich dataset where we have a large gap of $9.4\%$ and $4.7\%$ in $\kappa$ and $2.20\%$ and $2.60\%$ in AUC, in ViT and ResNet18, respectively. Furthermore, ZoomMIL is the best SOTA method in terms of generalizability, however MS-RGCN performs better by a large margin on source, Zurich, and Colorado datasets.

It is important to note that, for a fair comparison of our proposed model with the state-of-the-art, we tried various hyper-parameter settings for competing algorithms to make sure hyper-parameter setting did not contribute to inferior performance of the state-of-the-art algorithms (further details in Supplemental text; Section \textit{0.2}.).

\begin{figure*}[!t]
  \centering
  \includegraphics[width=\linewidth]{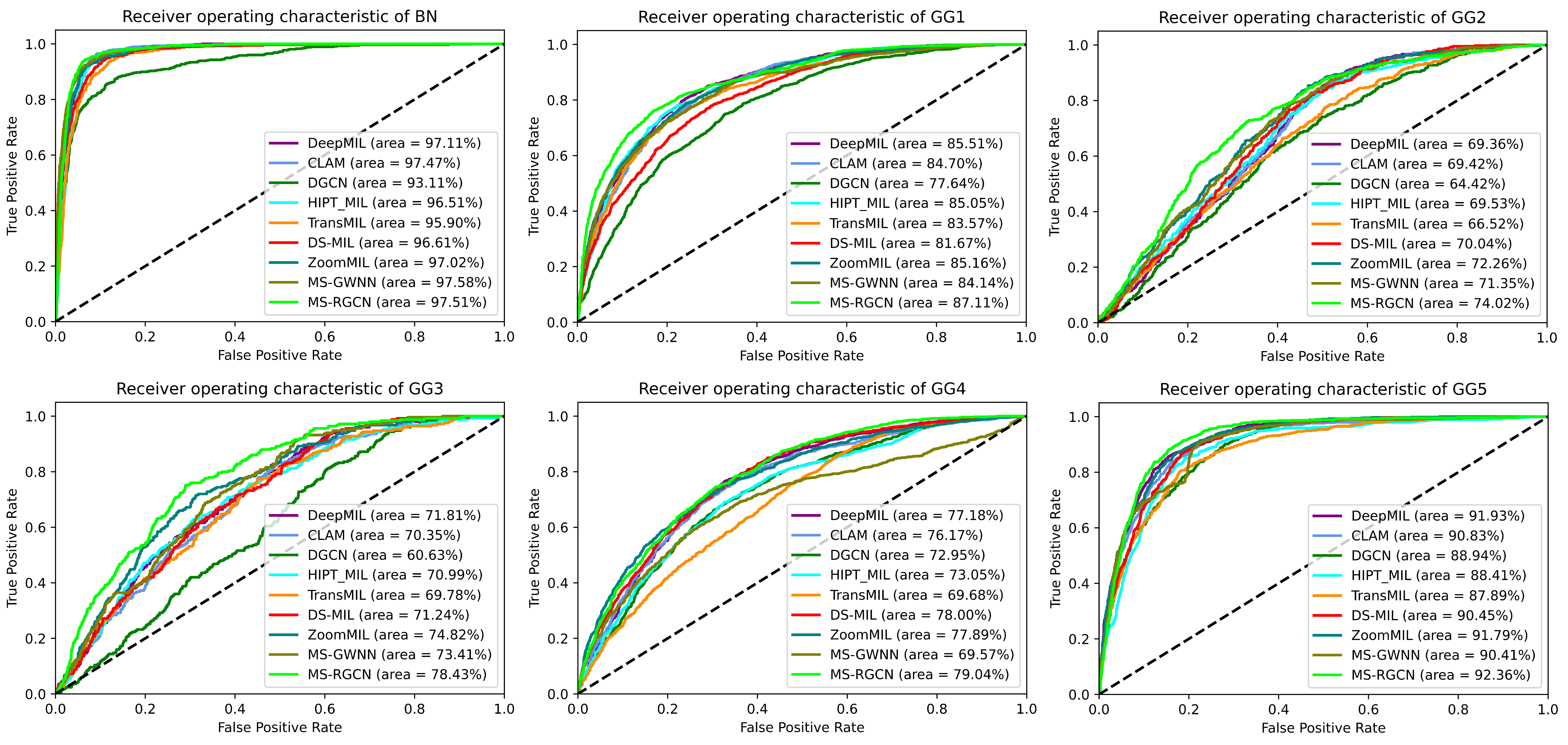}
  \caption{AUC performance of SOTA models on the Zurich (held-out) dataset based on each class across all of the folds. It can be seen that our model is the best in all of the GGs except benign which is the easiest task and our model is the second best with an insignificant difference of $0.07\%$.}
  \label{fig:auc}
\end{figure*}

\begin{figure*}[!t]
  \centering
  \includegraphics[width=\linewidth]{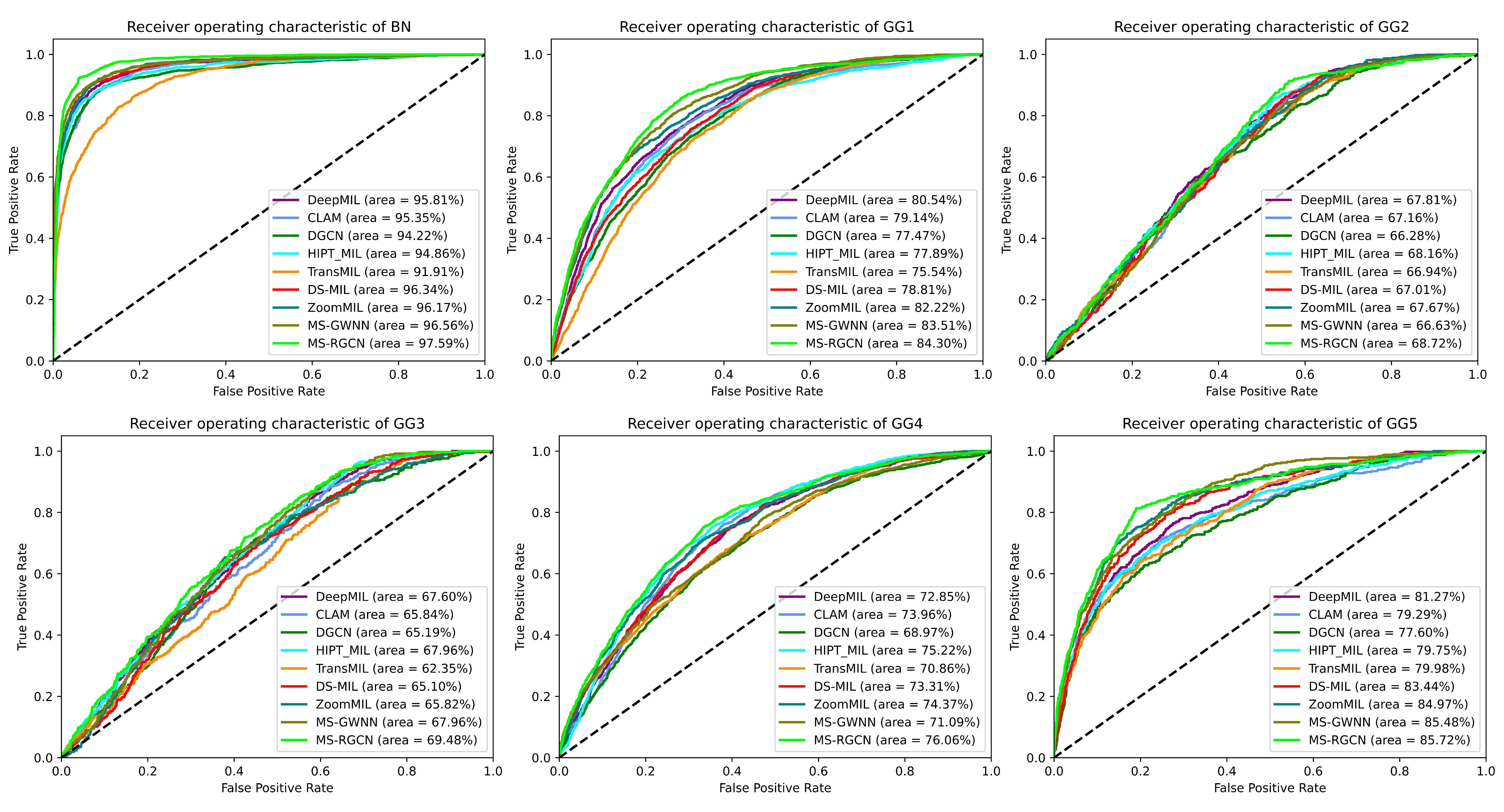}
  \caption{AUC performance of SOTA models on the Colorado dataset based on each class across all of the folds. Our model is the best in all of the GGs.}
  \label{fig:auc_colrado}
\end{figure*}

\begin{figure*}[!t]
  \centering
  \includegraphics[width=\linewidth]{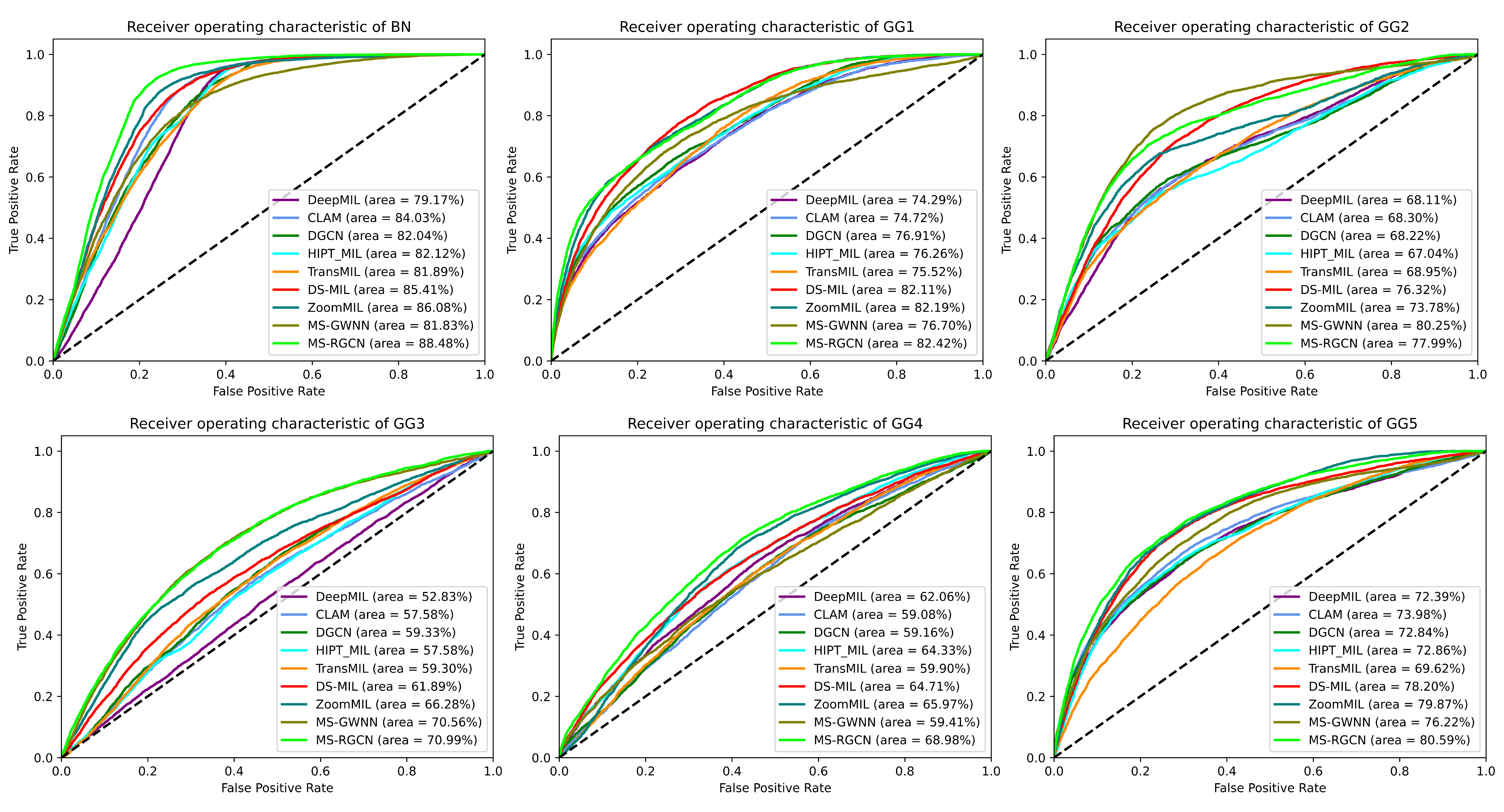}
  \caption{AUC performance of SOTA models on the Radboud dataset based on each class across all of the folds. Our model is the best in all of the GGs.}
  \label{fig:auc_panda}
\end{figure*}

Having better performance with ResNet18, we also compared the models with the ResNet18 feature extractor on the Colorado and PANDA (train on Karolinska and test on Radboud) datasets. Results in Table \ref{tab:sota_res} indicate that MS-RGCN outperforms the best SOTA models by $6.4\%$, and $2.7\%$ in $\kappa$ and $1.61\%$ and $2.50\%$ AUC, in Colorado and Radboud datasets, respectively. Thus, the proposed MS-RGCN is outperforming eight SOTA methods in various datasets and feature extractors. It is worth mentioning that the lower performance on the Radboud dataset from the PANDA challenge  in comparison to the reported results in that challenge \cite{bulten2022artificial} is due to the difference in the test datasets. Since the PANDA challenge \textit{test} dataset is not publicly available, we utilized the two training datasets that were available. However, as mentioned in their paper \cite{bulten2022artificial} and data website, due to the high number of slides in the training set they relied on pathology reports to extract Gleason scores associated with patients in the training data which could be noisy and less reliable in comparison to validation and test sets where the slides were annotated by a consensus of several pathologists.

\subsection{Class-based performance}
In Figure \ref{fig:auc}, the average performance of state-of-the-art models with ResNet18 feature extractor for each class of the Zurich dataset is shown. 
Our model is performing well across all of the classes, being the first in all of GG1-5 with a margin of 1.6\%, 1.76\%, 3.61\%, 1.15\%, and 0.43\% in one-vs-all AUC compared to the best SOTA model, respectively, and second in benign with negligible difference of $97.58\% - 97.51\% = 0.07\%$ after MS-GWNN. Moreover, it performs substantially better GG2 (GS3+4) and GG3 (GS4+3) classification tasks which are deemed to be more challenging where the model has to distinguish GP3 and GP4, by up to $9.60\%$ and $17.80\%$ in one-vs-all AUC, respectively.

In addition, the average performance of SOTA models with ResNet18 feature extractor for each class of the Colorado and Radboud datasets is shown in Figures \ref{fig:auc_colrado} and \ref{fig:auc_panda}. In the Colorado dataset, it can be seen that MS-RGCN outperforms SOTA models in all GG classification tasks. Improvement of the MS-RGCN in comparison to the best SOTA model in each GG is $1.03\%$, $0.79\%$, $0.56\%$, $1.52\%$, $0.84\%$, and $0.24\%$ in AUC. In the Radboud dataset, the proposed model outperforms SOTA models in all of the GGs except for GG2 where it is the second-best model after MS-GWNN.

\subsection{Visualization of the prediction heatmaps}

\begin{figure*}[!t]
  \centering
  \includegraphics[width=\linewidth]{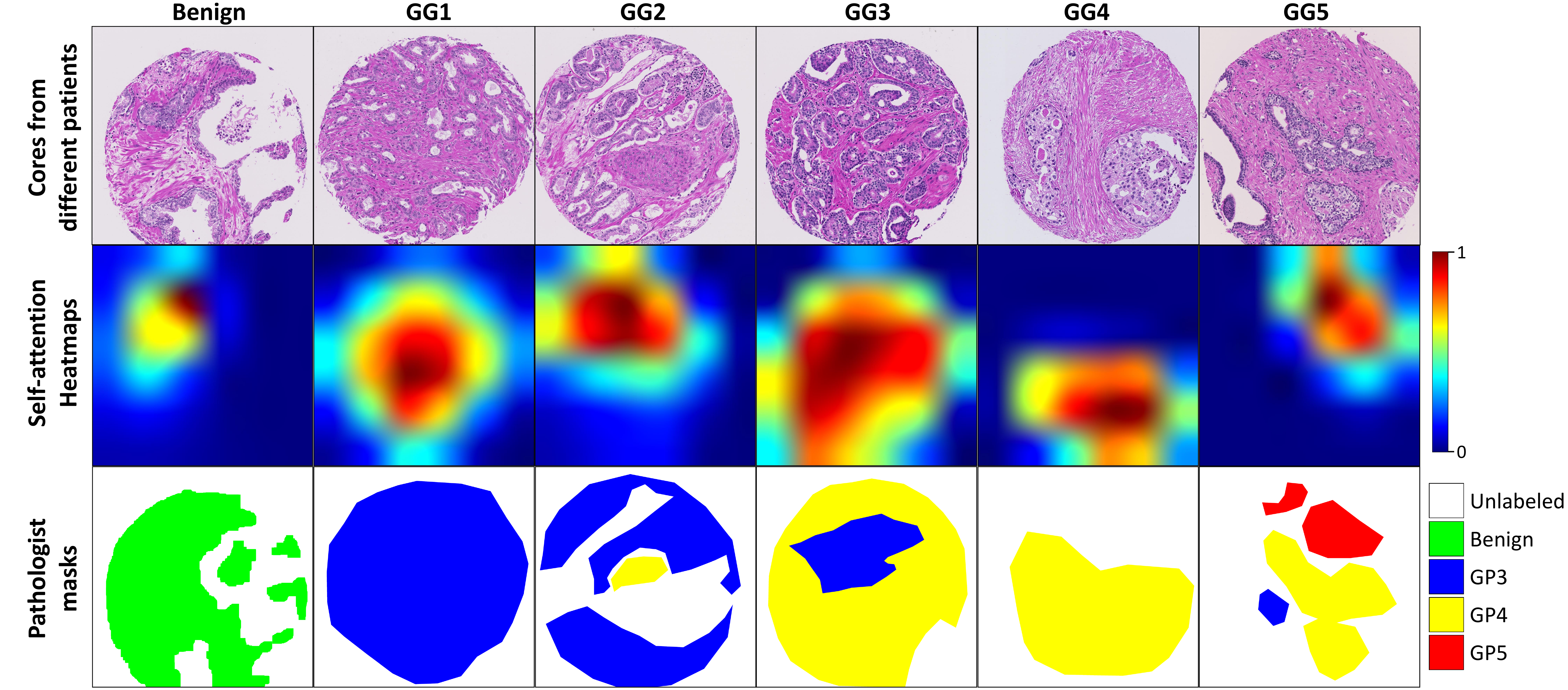}
  \caption{Self-attention heatmaps of the MS-RGCN on one sample from each class in the Zurich dataset. It can be seen that the model has focused on the important location for the classification, such as where the gland is, where the annotations are, and where the most important annotation is.}
  \label{fig:heatmap}
\end{figure*}

\begin{figure*}[!t]
  \centering
  \includegraphics[width=\linewidth]{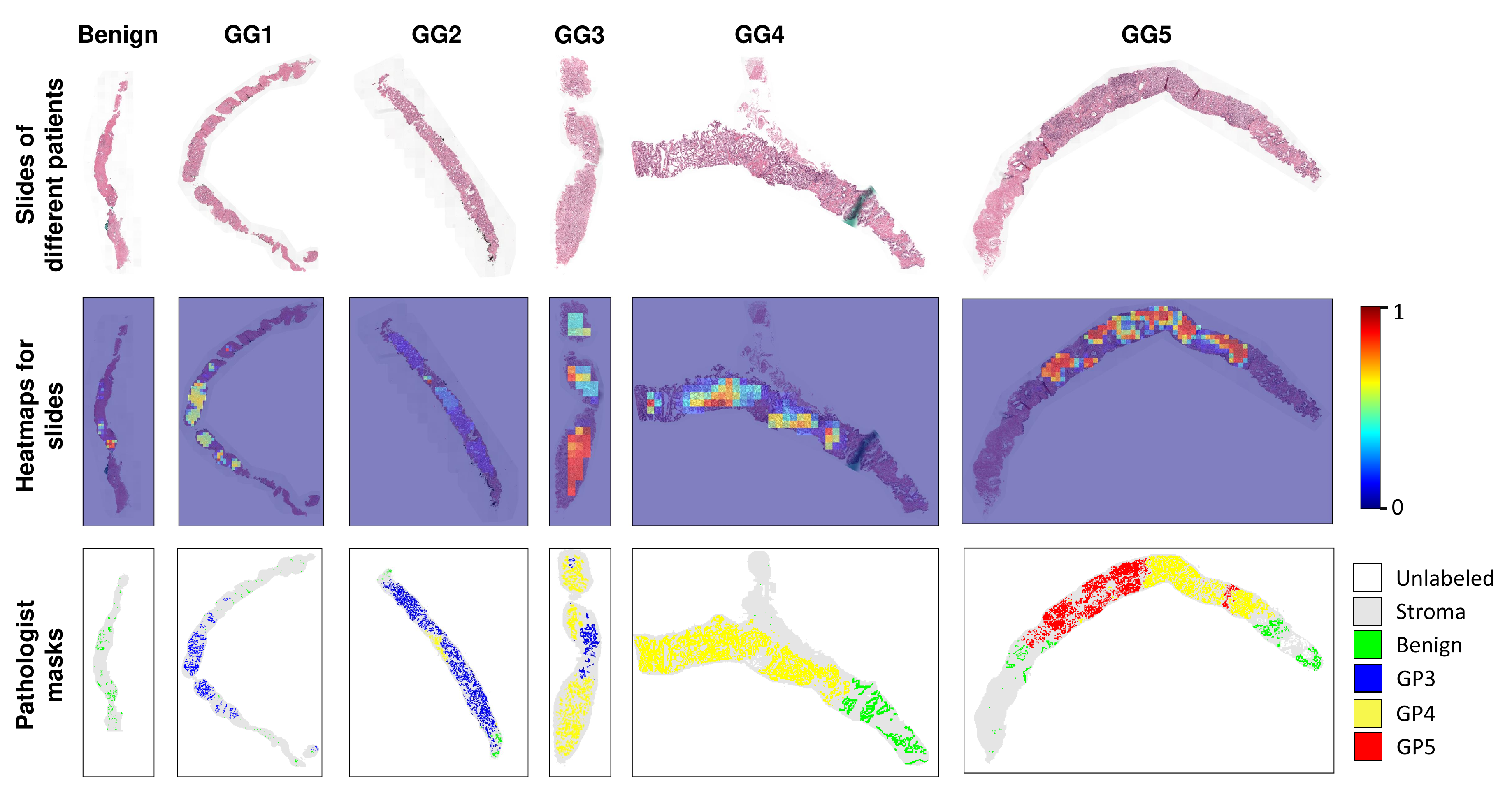}
  \caption{Self-attention heatmaps of the MS-RGCN on one sample from each class in the Radboud dataset. It can be seen that the model has focused on the important location for the classification, such as where the gland is, where the annotations are, and where the most important annotation is.}
  \label{fig:heatmap_panda}
\end{figure*}

Figure \ref{fig:heatmap} illustrates the attention heatmap of the model for the final prediction where red shows the highest attention and blue the lowest. We have plotted the attention for one sample per class and the model attention demonstrates the ability of the model to select the important regions for the final prediction. In the first column, we see a benign core, where the model has focused on the benign gland structure to identify the image as benign. In the second column, the model has focused on the whole core which contains GP3. In the third and fourth columns, the model has attended to both GP3 and GP4 regions to correctly classify the cores to GG2 (GS3+4) and GG3 (GS4+3). In the fifth column, attentions are only on the annotated regions (GP4) and the model has successfully ignored the unlabelled regions. Finally, in the last column, although we have GP3-5, the model has focused on GP5 to correctly predict the final GG.

In addition, Figure \ref{fig:heatmap_panda} shows the heatmaps on WSIs of the Radboud dataset. It can be observed that for the benign slide, the model has focused on the Benign section. In GG1, it ignored benign regions and only focused on the GP3 regions. In GG2 and GG3 slides, it focused more on the GP4 section which is more aggressive but also paid attention to the GP3 section to correctly classify them as GP3+GP4 (GG2) and GP4+GP3 (GG3). Finally, on the GG4 and GG5, attentions were on GP4 and GP4+GP5 which are the important components for predicting the grade group of the slide.

\subsection{Ablation study} \label{sec:ablation}

\begin{table*}[!t]
\centering
\caption{Ablation study on MIL methods with ViT feature extractor. We report the average and standard deviation of macro-averaged AUC and quadratic weighted $\kappa$ on 5-folds. The best performance is shown in {\bf bold}. 
}
\begin{tabular}{ccccc}
\hline
  \multirow{2}{*}{Method} & \multicolumn{2}{c}{{Vancouver dataset}} & \multicolumn{2}{c}{{Zurich dataset}} \\ \cline{2-5}
& kappa & AUC (\%) & kappa & AUC (\%) \\ \hline
{MS-RGCN-$5\times$} &  $0.543 \pm 0.069$ &$76.74 \pm 2.26$ & $0.709 \pm 0.018$& $80.52 \pm 0.73$ \\
{MS-RGCN-$10\times$} & $0.621 \pm 0.025$ &$81.12 \pm 1.86$ & $0.736 \pm 0.027$& $81.52 \pm 0.65$ \\
{MS-RGCN-$20\times$} & $0.662 \pm 0.036$ &$81.57 \pm 1.09$ & $0.716 \pm 0.027$& $81.08 \pm 1.57$ \\
{MS-RGCN-GE} & $0.703 \pm 0.046$ &$83.17 \pm 2.16$ & $0.745 \pm 0.019$& $83.06 \pm 1.02$ \\
{MS-RGCN-w/o-SE} & $0.686 \pm 0.036$ &$82.50 \pm 1.96$ & $0.750 \pm 0.030$& $82.66 \pm 1.73$ \\
{MS-GCN} & $0.619 \pm 0.038$ &$80.48 \pm 2.24$ & $0.754 \pm 0.014$& $82.50 \pm 1.03$ \\
{MS-RGCN-4ReLU} & $0.641 \pm 0.061$ &$82.29 \pm 2.88$ & $0.753 \pm 0.018$& $82.57 \pm 0.83$ \\ \hline
{{MS-RGCN (ours)}} & $\mathbf{0.712 \pm 0.061}$ &$\mathbf{83.97 \pm 1.64}$ & $\mathbf{0.765 \pm 0.020}$& $\mathbf{83.12 \pm 0.60}$ \\ \hline
\end{tabular}%
\label{tab:ablation}
\end{table*}

\begin{table*}[!t]
\centering
\caption{Ablation study on MIL methods with ResNet18 feature extractor. We report the average and standard deviation of macro-averaged AUC and quadratic weighted $\kappa$ on 5-folds. The best performance is shown in {\bf bold}. 
}
\resizebox{\textwidth}{!}{%
\begin{tabular}{ccccccccc}
\hline
\multirow{2}{*}{Method} & \multicolumn{2}{c}{{Vancouver dataset}} & \multicolumn{2}{c}{{Zurich dataset}} & \multicolumn{2}{c}{{Colorado dataset}} & \multicolumn{2}{c}{{Radboud dataset}} \\ \cline{2-9}
 & kappa & AUC (\%) & kappa & AUC (\%) & kappa & AUC (\%) & kappa & AUC (\%) \\ \hline
{MS-RGCN-$5\times$} & $0.623 \pm 0.028$ &$80.62 \pm 0.31$ & $0.719 \pm 0.020$ & $82.75 \pm 1.44$ & $0.604 \pm 0.036$& $78.11 \pm 0.59$ & $0.481 \pm 0.038$ & $77.37 \pm 1.39$ \\
{MS-RGCN-$10\times$} & $0.716 \pm 0.022$ &$83.35 \pm 1.03$ & $0.758 \pm 0.047$ & $83.66 \pm 2.29$ & $0.555 \pm 0.103$& $79.25 \pm 1.39$ & $0.545 \pm 0.044$ & $77.79 \pm 1.39$ \\
{MS-RGCN-$20\times$} & $0.704 \pm 0.040$ &$83.01 \pm 2.04$ & $0.721 \pm 0.047$ & $81.53 \pm 1.62$ & $0.588 \pm 0.083$& $76.40 \pm 4.02$ & $0.386 \pm 0.061$ & $68.84 \pm 3.72
$ \\
{MS-RGCN-GE} & $0.691 \pm 0.034$ &$83.79 \pm 1.25$ & $0.757 \pm 0.015$& $83.99 \pm 1.52$ & $0.656 \pm 0.017$ & $80.16 \pm 1.57$ & \multicolumn{2}{c}{{out of memory error}} \\
{MS-RGCN-w/o-SE} & $0.708 \pm 0.034$ &$84.20 \pm 0.82$ & $0.756 \pm 0.042$ & $84.31 \pm 1.40$ & $0.649 \pm 0.024$ & $78.93 \pm 1.59$ & $0.565 \pm 0.019$ & $77.30 \pm 3.42$ \\
{MS-GCN} & $0.694 \pm 0.037$ &$83.49 \pm 1.84$ & $0.754 \pm 0.061$& $84.08 \pm 2.09$ & $0.657 \pm 0.036$& $78.88 \pm 1.40$ & $0.560 \pm 0.030$ & $77.35 \pm 3.28$ \\
{MS-RGCN-4ReLU} & $0.718 \pm 0.044$ &$84.38 \pm 1.77$ & $0.779 \pm 0.039$& $84.16 \pm 1.99$ & $0.649 \pm 0.047$& $79.83 \pm 1.16$ & $0.571 \pm 0.033$ & $78.18 \pm 2.67$ \\ \hline
{{MS-RGCN (ours)}} & $\mathbf{0.727 \pm 0.028}$ &$\mathbf{85.18 \pm 1.45}$ & $\mathbf{0.781 \pm 0.013}$ & $\mathbf{85.24 \pm 1.07}$ & $\mathbf{0.668 \pm 0.031}$& $\mathbf{80.86 \pm 0.89}$ & $\mathbf{0.577 \pm 0.025}$& $\mathbf{78.84 \pm 2.73}$ \\ \hline
\end{tabular}%
}
\label{tab:ablation_res}
\end{table*}

For the ablation study, we experimented with seven models in addition to our model. MS-RGCN-$5\times$, MS-RGCN-$10\times$, and MS-RGCN-$20\times$ are the proposed MS-RGCN in single $5\times$, $10\times$, and $20\times$ magnifications. MS-RGCN-GE is the ablation study on graph edge definitions where there is an edge between two patches in one magnification ($E_n$) if their distance is less than half of the maximum distance, similar to the DGCN \cite{zhao2020predicting} graph definition. MS-RGCN without Scale Edges (MS-RGCN-w/o-SE) is obtained by removing the across-magnification edges in step 2 and only using steps 1, 3, and 4 in Figure \ref{fig:fig-graph}. Multi-Scale Graph Convolution Network (MS-GCN) is the same model as MS-RGCN but a homogenous graph where all edges and nodes are of the same type and the same message-passing network is used in all of the connections. In other words, MS-GCN does not have the relational and heterogeneous features that the proposed MS-RGCN has. MS-RGCN-4ReLU is the MS-RGCN model but with ReLU activation functions in the first step of the model where we did not use activation functions to prevent adding non-linearity and changing the features while we are trying to make them robust by seeing neighbors in a distance of 2 nodes in the graph.

As it can be seen in Tables \ref{tab:ablation} and \ref{tab:ablation_res}, MS-RGCN is substantially better than single magnification models, showing the effectiveness of multi-scale learning of the model. MS-RGCN-GE demonstrates that our locally defined edge is better compared to more global ones, as in DGCN. MS-RGCN-w/o-SE shows the importance of the second step in the model where we pass information across different scales. MS-GCN is very close to single magnification results, especially if we look at the $\kappa$ values. This shows the importance of using a relational graph and edge types in order to not mix information from different scales and get the highest possible performance with multi-scale information. Finally, MS-RGCN-4ReLU shows that seeing a larger area without changing the embeddings with non-linear activation functions is a better choice compared to having non-linear activation functions.

\subsection{Inference time and number of parameters}
Table \ref{tab:time} illustrates the inference time and the number of parameters for each MIL model. It can be observed that multi-scale methods have more parameters and at the same time gain better performance since they mimic pathologists' analysis better. For instance, ZoomMIL, the best SOTA model on Zurich and Colorado datasets, has the highest number of parameters (3.17 million) and, similarly, our proposed MS-RGCN has more parameters but also has superior performance compared to other techniques. 
In addition, our proposed method and CLAM-SB have the lowest inference time of 17.2 milliseconds in TMAs and our inference time is close to the best model in WSIs. The models are trained and tested using an NVIDIA\textsuperscript{®} Tesla\textsuperscript{®} V100 GPU with 16GB memory.

\begin{table}[!t]
\centering
\caption{Comparing Inference Time (IT) for TMA (Zurich dataset) and WSI (Radboud dataset) in milliseconds and number of parameters in millions with state-of-the-art MIL methods. The best performance is shown in {\bf bold}.
}
\begin{tabular}{cccc}
\hline
Method & IT (TMA) & IT (WSI) & Number of parameters \\ \hline
DeepMIL & $ 17.6 $ & $ \mathbf{83.3} $ & $ \mathbf{0.13\times10^6} $ \\
{CLAM-SB } & $ \mathbf{17.2} $ & $ 88.2 $ & $ 0.53\times10^6 $  \\
DGCN & $21.4$ & $ 89.3 $ & $0.66\times10^6$ \\
{HIPT-MIL } & $20.4$ & $ 94.2 $ & $0.66\times10^6$ \\
{TransMIL } & $27.3$ & $ 90.4 $ & $2.41\times10^6$ \\
{DS-MIL } & $17.6$ & $ 86.6 $ & $1.22\times10^6$ \\
{ZoomMIL } & $20.2$ & $ 86.9 $ & $3.17\times10^6$ \\
{MS-GWNN } & $17.4$ & $ 86.4 $ & $0.21\times10^6$ \\ \hline
{MS-RGCN (ours)} & $ \mathbf{17.2} $ & $ 86.5 $ & $4.24\times10^6$ \\ \hline
\end{tabular}%
\label{tab:time}
\end{table}

\section{Conclusion}
We believe our work is an important step towards improving GCN performance on different types of information for multiple instance learning with the help of RGCNs and different edge types.
To the best of our knowledge, this is the first paper that uses GCNs in multi-scale analysis of histopathology images to their fullest potential with early-fusion by using a combination of different edge types between neighboring patches and patches at different scales. This is helpful in handling different feature types in MIL problems.
The proposed method outperforms SOTA on the source and held-out datasets and is capable of classifying complex classes substantially better.
We developed this model for PCa which is a heterogeneous disease with a diverse range of histological patterns which make it challenging for diagnosis and prognosis. PCa remains the third-leading cause of cancer death in men, along with prostate-specific antigen, histopathology analysis of biopsy samples is a crucial step in deciding patient treatment \cite{litwin2017diagnosis}.

The limitation of this work is that in the end, attention-based mechanisms are required for image-level prediction in our model. Future directions include dedicating a node (as an output) to combine information from different levels of RGCNs; this would be similar to having skip connections for the final prediction.
We would like to point out that the idea of handling different embedding spaces using RGCNs could motivate multi-modal problems, such as Dwivedi and Shao \etal work \cite{dwivedi2022multi, shao2020improving}.

\section{Data and Code Availability}
Majority of the data associated with this manuscript is publicly available as follows:
\\(a) Vancouver dataset as part of the Gleason 2019 Challenge \href{https://tinyurl.com/VPCdataset}{https://tinyurl.com/VPCdataset},
\\(b) Zurich dataset: \href{https://tinyurl.com/Zurichdataset}{https://tinyurl.com/Zurichdataset},
\\(c) Karolinska and Radboud datasets from PANDA challenge: \href{https://tinyurl.com/PANDAdataset}{https://tinyurl.com/PANDAdataset}.
The code associated with this work will be available upon publication.


\section{Acknowledgement}
This work was supported by the Canadian Institute of Health Research (CIHR) [Grant \# 450849] and Michael Smith Health Research Scholar Award for AB.

\bibliographystyle{unsrt}  
\bibliography{references}  

\end{document}